\documentclass[11pt]{article}

\usepackage[final]{acl}

\usepackage{times}
\usepackage{latexsym}
\usepackage{booktabs}

\usepackage[T1]{fontenc}

\usepackage[utf8]{inputenc}

\usepackage{microtype}
\usepackage{subcaption}
\usepackage{inconsolata}
\usepackage{graphicx}

\title{No Text Needed: Forecasting MT Quality and Inequity from Fertility and Metadata}

\begin{document}

\author{%
  Jessica M.~Lundin \\
  Institute for Disease Modeling\\
  Gates Foundation \\
  \And
  Ada Zhang \\
  University of San Francisco \\
  \AND
  David Ifeoluwa Adelani \\
  Mila, McGill University \& \\
  Canada CIFAR AI Chair \\
  \And
  Cody Carroll \\
  University of San Francisco
}

\maketitle

\begin{abstract}
We show that translation quality can be predicted with surprising accuracy from token-level statistics and linguistic metatdata alone, \textit{without inspecting the translated text}. Using only a handful of features: token fertility ratios, token counts, and basic linguistic metadata (language family, script, and region), we can forecast ChrF scores for GPT-4o translations across languages in the FLORES-200 benchmark. Gradient boosting models achieve favorable performance ($R^{2}=0.66$ for XX$\rightarrow$English and $R^{2}=0.72$ for English$\rightarrow$XX).  Feature importance analyses reveal that typological factors dominate predictions into English, while fertility plays a larger role for translations into diverse target languages, and the importance of fertility varies by model. We are not proposing methodology for quality estimation, rather these findings suggest explainability of translation quality shaped by both token-level fertility and broader linguistic typology, offering insight for multilingual evaluation and quality estimation.  Our findings reveal systematic performance disparities across language families and regions, with implications for fairness and equity in multilingual NLP systems.
\end{abstract}

\section{Introduction}

Machine translation (MT) quality evaluation has evolved from early rule-based and statistical methods to large-scale neural models. Traditional evaluation metrics such as BLEU \cite{papineni2002bleu} and METEOR \cite{banerjee2005meteor} have been widely used but have been criticized for their limited sensitivity to linguistic diversity and their reliance on surface-level n-gram matches. More recent alternatives, such as ChrF \cite{popovic2015chrf} and ChrF++ \cite{popovic2017chrfpp}, leverage character-level representations to better capture morphologically rich languages, and have shown strong correlation with human judgments in multilingual settings.  The difference between quality estimation and this work is that we are not proposing a method for quality estimation, but rather, seek to understand systematic cross-linguistic patterns in model behavior.  

\begin{figure}[h]
    \centering
    \includegraphics[width=0.48\textwidth]{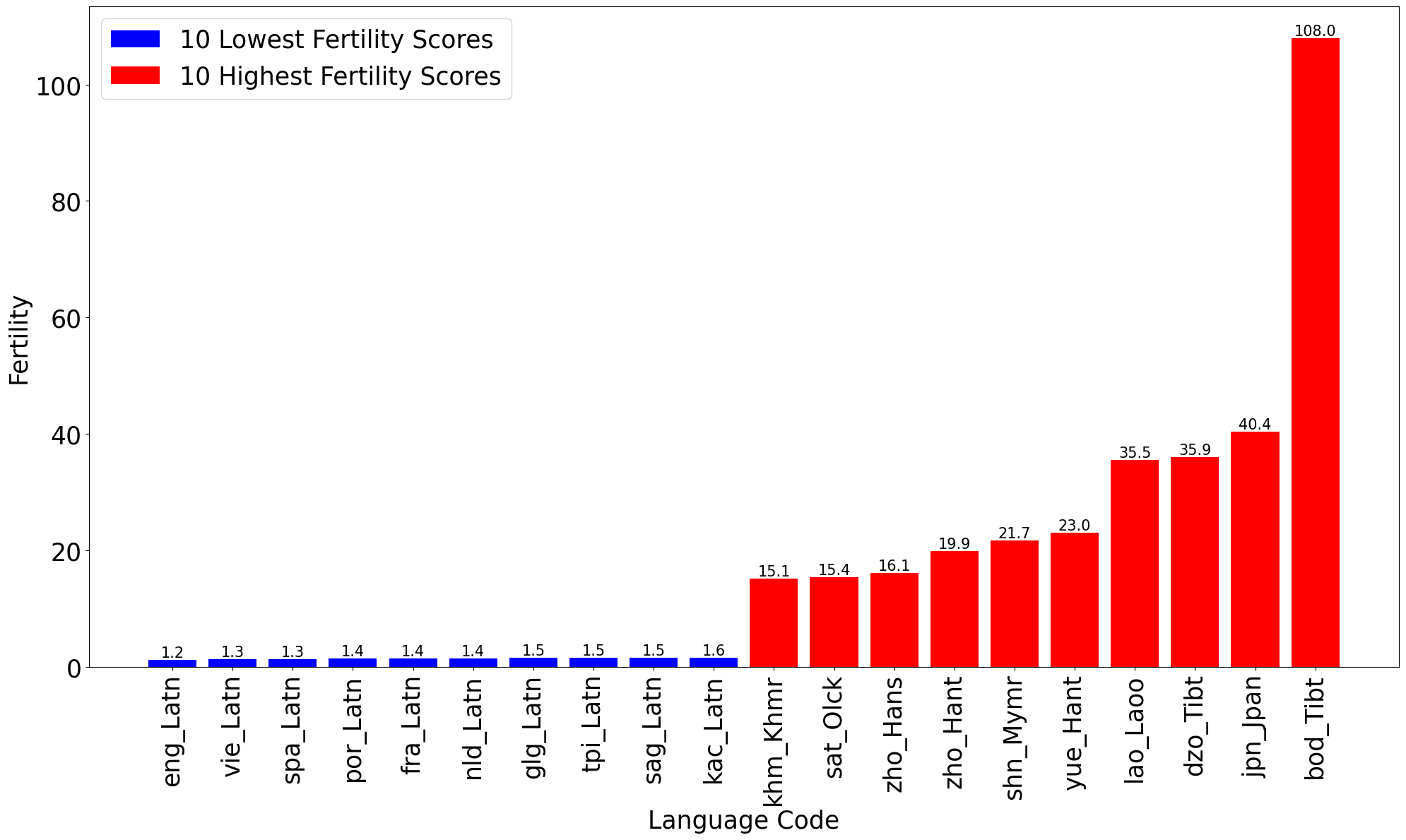}
    \caption{Languages with highest and lowest fertility for FLORES-200 dataset.  Latin scripts tend to have lower fertility values.}
    \label{fig:flores-fertility}
\end{figure}

One central factor in MT quality is fertility, originally formalized in IBM Models for statistical machine translation \cite{brown1993mathematics}. Fertility refers to how many target tokens are generated per source word, and imbalances in fertility often lead to errors such as under-translation or over-translation. Figure \ref{fig:flores-fertility} shows the language codes with highest and lowest fertility values for FLORES-200.  While fertility has historically been studied in statistical MT, its role in neural MT evaluation remains underexplored. Recent work in large-scale multilingual evaluation, such as MEGA \cite{ahuja2023mega}, has underscored the importance of tokenization and representation choices in shaping model performance across diverse languages, suggesting that fertility may still play a critical role.

At the same time, linguistic metadata such as language family, script type, and geographic region have been shown to influence translation performance. For example, scaling efforts such as NLLB \cite{costa2024nllb}, SIB-200~\cite{adelani-etal-2024-sib} and mSTEB~\cite{beyene2025msteb} demonstrate that typological and resource disparities across languages lead to systematic variation in model accuracy. Similarly, the WMT Quality Estimation shared tasks \cite{specia2020wmtqe} have highlighted the value of incorporating language-level features into predictive models of translation quality.

These findings point to the need for a systematic investigation of how fertility and linguistic metadata interact with translation quality.  Rather than building a runtime quality estimator, our research question: What factors explain quality variation across 200 languages?  By modeling these features explicitly, we find interpretable patterns.

\section{Methods}

We developed 5 regression models (Linear, Lasso, MLP, Random Forest, XGBoost) to predict ChrF (Character n-gram F-score) translation quality scores for GPT-4o translations in the FLORES-200 benchmark dataset \cite{nllb-24}. 

For data, we utilize the LLM text translations of the FLORES-200 benchmark and annotated features released by mSTEB~\cite{beyene2025msteb}. Our analysis covered two translation directions: multilingual-to-English (XX→English) and English-to-multilingual (English→XX) across 200 languages. 

Table \ref{tab:features} shows features used to fit the ChrF regression.  We extracted both linguistic and text-level features from the translation pairs. Text-level features included token counts for both source and target texts using the ``\texttt{o200kbase}'' tokenizer, as well as fertility ratios
(tokens per word) for source and target texts.  Language-level metadata include language (ISO language codes, script (29 scripts), Joshi class (0-5), language family, and geographic region.  
We compared the fit of multiple regression approaches: \textbf{Linear (OLS, Lasso)}, \textbf{Tree ensembles (Random Forest, XGBoost)}, and \textbf{multi-layer perceptron (MLP)}, using metrics R$^2$, RMSE, MAE.  For reproducibility, we used 20\% hold out and values from hyperparameter grid search are shown in Table \ref{tab:regression_hyperparameters}. 

Feature importances were extracted from the trained Random Forest and XGBoost models to identify which variables most strongly influence translation quality predictions. For Random Forest models, importances are calculated using mean decrease in impurity (Gini importance).  During tree construction, each feature's contribution to reducing variance at split nodes is tracked and averaged across all 300 trees in the forest. Features that consistently produce larger reductions in the residual sum of squares when used for splitting receive higher importance scores. For XGBoost models, we use the default gain-based importance metric, which measures the average improvement in prediction accuracy (reduction in loss function) contributed by each feature across all trees.  

Marginal averages are calculated by first training Random Forest and XGBoost regression models on the training set using all features (categorical language features and numeric tokenization features). After training, the models generate predictions on the held-out test set. For each categorical feature (Region, Family, Script, Joshi Class, Language Code), we group the test set observations by their categorical values and compute the mean predicted score within each group. For example, to determine the marginal average score for "Africa," we average all predicted translation quality scores for test samples where the region is Africa, marginalizing over all other feature values (different language families, scripts, fertility rates, etc.). Both predicted means and actual means are calculated for each category to assess model fit.

\begin{table*}[!htbp]
\centering

\resizebox{\textwidth}{!}{
\footnotesize
\begin{tabular}{p{1.5cm}p{2.5cm}p{10cm}}
\textbf{Feature} & \textbf{Example} & \textbf{Description} \\
\midrule
Joshi Class & 0-5 & Joshi class labels \cite{joshi2021statefatelinguisticdiversity}, categorizing languages by resource availability and computational support.  These are imputed for the FLORES-200 dataset. \\
\midrule
\addlinespace
Region & Africa, Europe, Americas & 9 geographic regions where the language is primarily spoken, used to capture regional linguistic patterns \\
\midrule
\addlinespace
Family & Austronesian, Afro-Asiatic, Indo-European & Linguistic family classification, grouping languages by common ancestral origins and structural similarities \\
\midrule
\addlinespace
Script & Arab, Latn, Cyrl, Deva & 29 scripts used (Arabic, Latin, Cyrillic, Devanagari, etc.) \\
\midrule
\addlinespace
Code & ace, afr, amh, ara & ISO language code identifier, uniquely identifying each language \\

\midrule
\addlinespace
Reference Fertility & 2.5, 3.2, 4.1 & Average number of tokens per word in reference (human) translations, measuring morphological complexity \\

\midrule
\addlinespace
Candidate Fertility & 2.3, 3.0, 3.8 & Average number of tokens per word in LLM-generated translations, measuring model's tokenization efficiency \\

\midrule
\addlinespace
Reference Tokens & 54, 101, 160 & Total number of tokens in the reference translation for a given text \\

\midrule
\addlinespace

Candidate Tokens & 48, 78, 200 & Total number of tokens in the LLM-generated translation for a given text \\
\hline
\end{tabular}
}

\caption{Features used in the ChrF predictive analysis.}
\label{tab:features}
\end{table*}

\begin{table}[h]
\centering

\scriptsize
\begin{tabular}{llp{4cm}}
\hline
\textbf{Model} & \textbf{Direction} & \textbf{Hyperparameters} \\
\hline
Linear Regression & both & tol=1e-06 \\
\hline
LassoCV & En$\rightarrow$XX & cv=5, $\alpha$=0.0347 \\
 & XX$\rightarrow$En & cv=5, $\alpha$=0.0191 \\
\hline
Random & both & \begin{tabular}[c]{@{}l@{}}n\_est=300, depth=None, feat='sqrt',\\ split=2, leaf=5, seed=42\end{tabular} \\
Forest & & \\
\hline
XGBoost & En$\rightarrow$XX & \begin{tabular}[c]{@{}l@{}}n\_est=300, depth=7, lr=0.1, sub=0.6,\\ colsamp=1.0, child\_wt=5, \end{tabular} \\
 & XX$\rightarrow$En & \begin{tabular}[c]{@{}l@{}}n\_est=300, depth=10, lr=0.05, sub=0.8,\\ colsamp=0.8, child\_wt=1, \end{tabular} \\
\hline
MLP & En$\rightarrow$XX & layers=(128,64), act=tanh, ep=50, batch=64 \\
 & XX$\rightarrow$En & layers=(256,128,64), act=relu, ep=50, batch=32 \\
\hline
\end{tabular}

\caption{Sklearn \cite{scikit-learn} hyperparameters for regression models predicting translation quality}
\label{tab:regression_hyperparameters}

\end{table}

\section{Results and Discussion}

\subsection{Model Performance Comparison}

Table~\ref{tab:model_performance} demonstrates a clear performance hierarchy across all model types and translation directions. The substantial performance gap between linear models (R² $\approx$ 0.25-0.31) and tree-based models (R² $\approx$ 0.66-0.72) indicates strong non-linear relationships in the data that simple linear combinations cannot capture effectively. XGBoost achieves the highest performance in both directions: English-to-XX and XX-to-English respectively had R² values of 0.719 and 0.663, while Random Forest achieved values of 0.701 and 0.588. 

Neural networks show moderate performance (0.586 train R² and 0.684 test R²) but remain substantially below ensemble methods, while linear approaches (Linear and Lasso Regression) perform poorly across both directions with nearly identical results. The consistent superiority of XGBoost over Random Forest suggests that gradient boosting is particularly well-suited for capturing the complex interactions between language-level metadata and translation quality, with performance gaps of 0.018 R² for English-to-XX and 0.075 R² for XX-to-English translation.

\begin{table*}[!htbp]
\centering

\begin{tabular}{|l|ccc|ccc|}
\toprule
\hline
\textbf{Model} & \multicolumn{3}{|c|}{\textbf{XX→English}} & \multicolumn{3}{|c|}{\textbf{English→XX}} \\
\hline
 & \textbf{R²} & \textbf{RMSE} & \textbf{MAE} & \textbf{R²} & \textbf{RMSE} & \textbf{MAE} \\
 \hline
\midrule
Linear Regression & 0.25 & 16.58 & 13.50 & 0.31 & 16.19 & 13.01 \\
Lasso Regression & 0.25 & 16.58 & 13.51 & 0.31 & 16.19 & 13.01 \\
MLP & 0.59 & 12.34 & 9.71 & 0.68 & 10.97 & 8.33 \\
Random Forest & 0.59 & 12.31 & 9.73 & 0.70 & 10.68 & 8.11 \\
XGBoost & 0.66 & 11.14 & 8.71 & 0.72 & 10.36 & 7.88 \\
\hline
\bottomrule
\end{tabular}
\caption{Model Performance Comparison across Translation Directions}
\label{tab:model_performance}

\end{table*}

\begin{figure*}[htbp]
\centering
\begin{subfigure}[b]{0.48\textwidth}
    \centering
    \includegraphics[width=\textwidth]{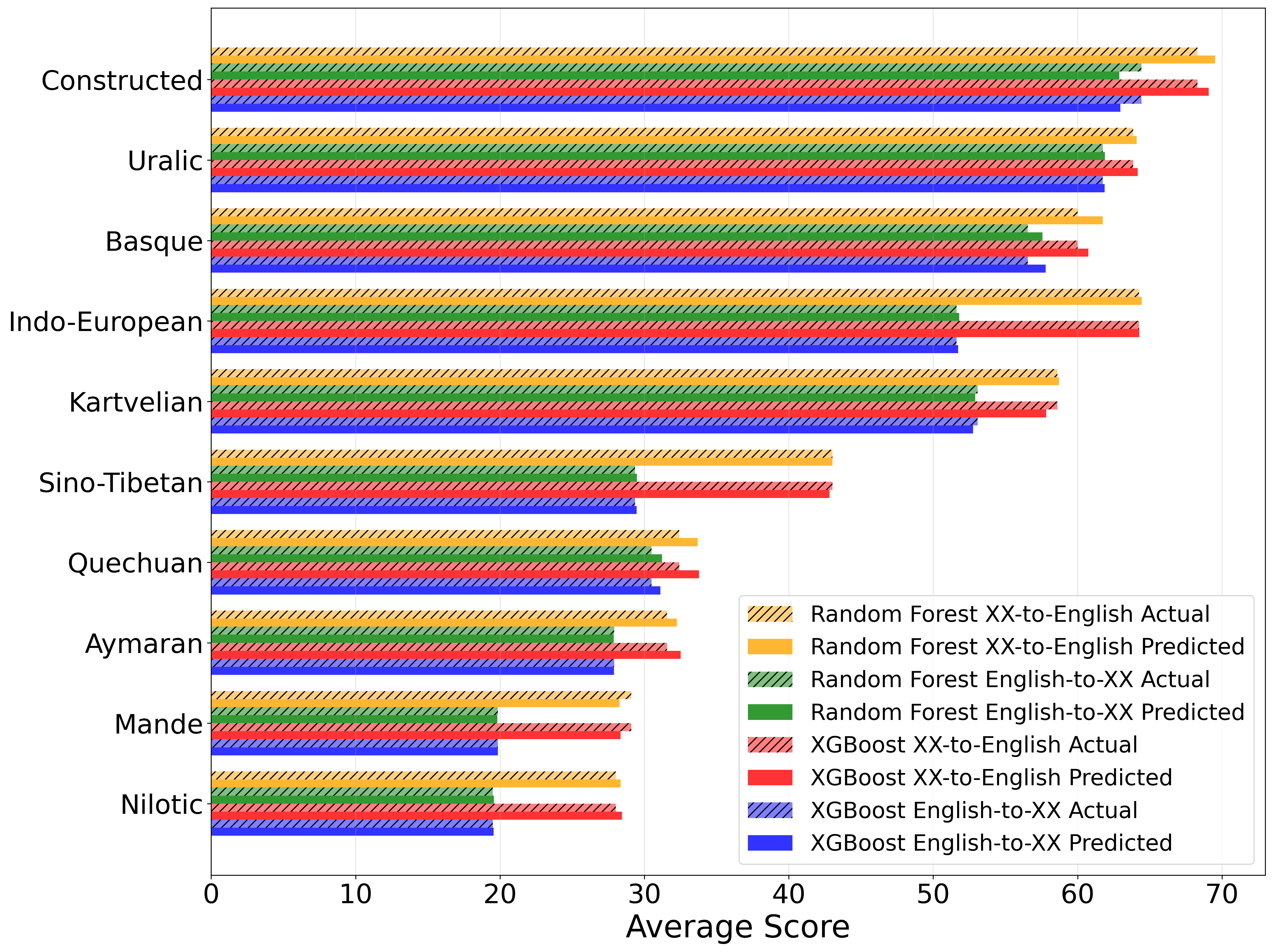}
    \caption{Language Family}
    \label{fig:family_comparison}
\end{subfigure}
\hfill
\begin{subfigure}[b]{0.48\textwidth}
    \centering
    \includegraphics[width=\textwidth]{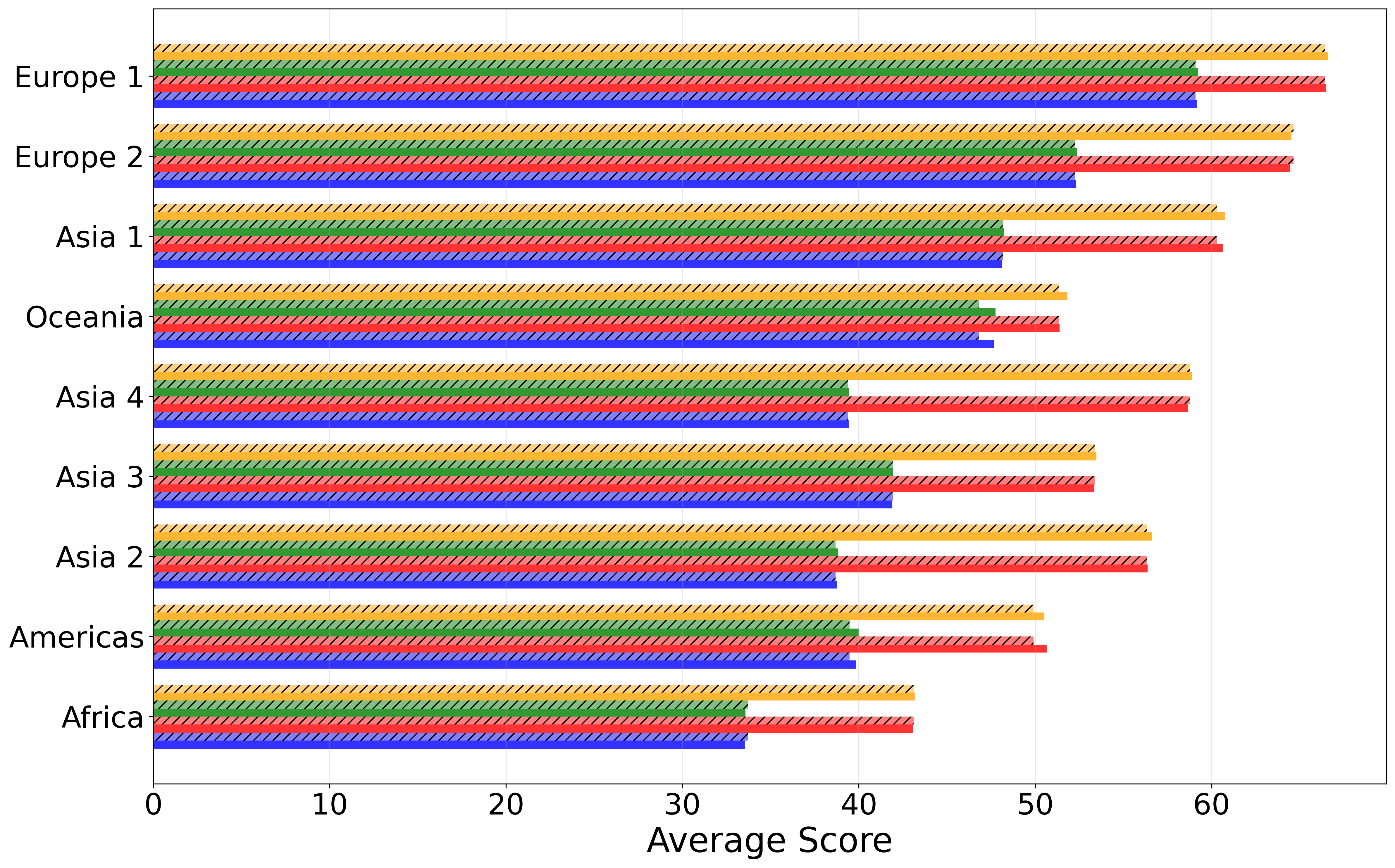}
    \caption{Region}
    \label{fig:region_comparison}
\end{subfigure}

\vspace{1em}

\begin{subfigure}[b]{0.48\textwidth}
    \centering
    \includegraphics[width=\textwidth]{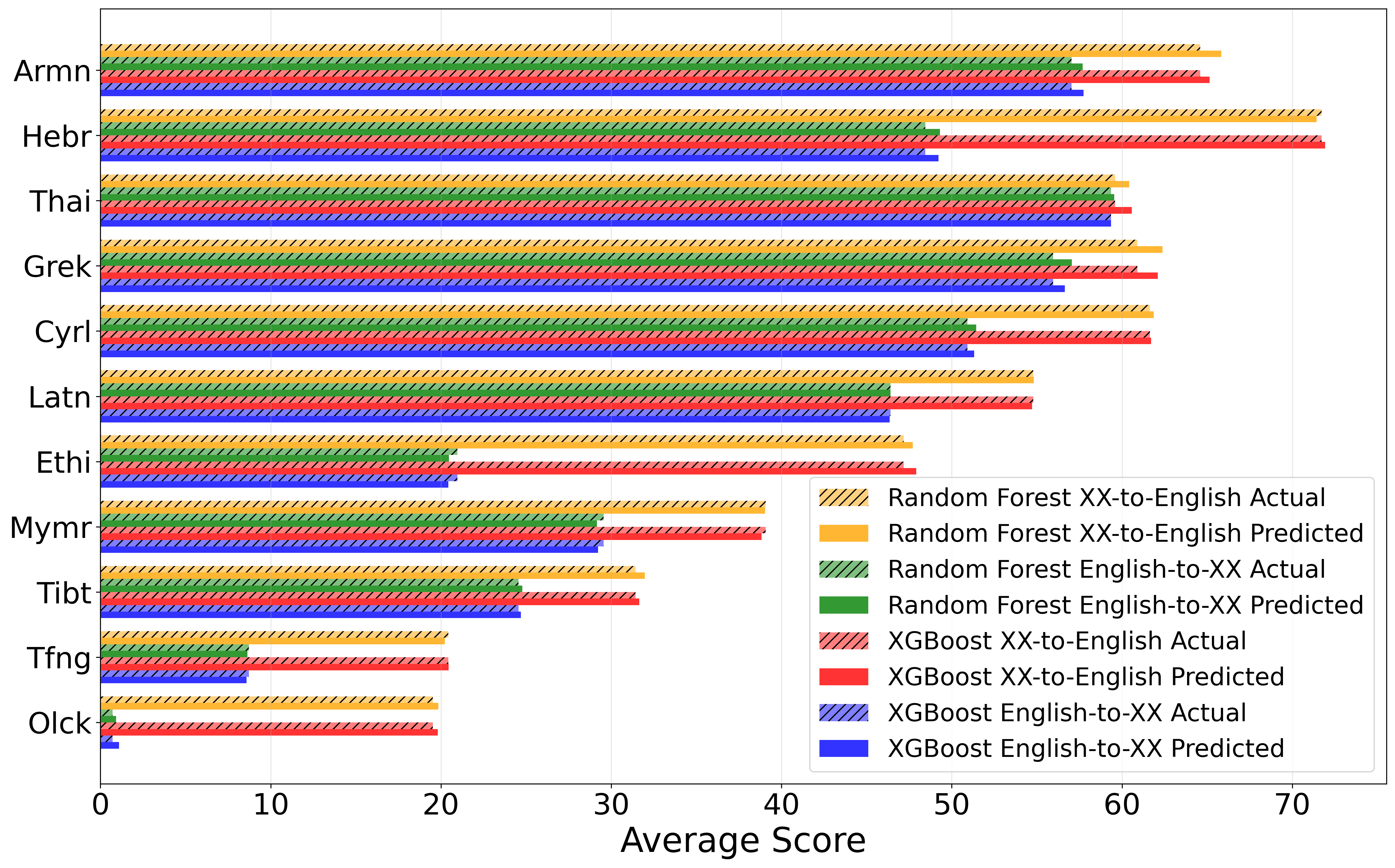}
    \caption{Script}
    \label{fig:script_comparison}
\end{subfigure}
\caption{Marginal Average Translation Quality Scores by Categorical Features. Each panel shows XGBoost vs Random Forest performance across both English-to-XX and XX-to-English translation directions, displaying both predicted and actual scores. Panel (a) examines the top and bottom 5 language families, demonstrating the substantial performance gap between high-resource families like Constructed and Indo-European languages versus low-resource families such as Mande and Nilotic. Panel (b) compares performance across geographic regions, revealing systematic differences where European languages consistently have the highest performance. Panel (c) shows performance by script, with the top 6 (Latn is 6th) and bottom 5 scripts.  Across panels, predicted scores (solid bars) align closely with actual scores (hatched bars), indicating good model calibration, and XGBoost and Random Forest show remarkably similar performance patterns, suggesting that the underlying linguistic patterns are robust across different ensemble methods.}
\label{fig:model_comparisons}
\end{figure*}

Figure \ref{fig:model_comparisons} presents a comparison of marginal average translation quality scores across three key categorical features: geographic region, language family, and script. The analysis reveals systematic performance patterns that are remarkably consistent between XGBoost and Random Forest models across both translation directions, indicating that these linguistic biases are inherent to the underlying data rather than model-specific artifacts. Language family effects (\ref{fig:family_comparison}) show dramatic variation between resource levels, where constructed languages like Esperanto and high-resource families like Indo-European score 15-20 points higher than low-resource families such as Niger-Congo and Austronesian. Regional disparities (\ref{fig:region_comparison}) demonstrate clear geographic variation, with European languages achieving scores of 55-65 compared to 35-45 for African languages. Effects from the script type (\ref{fig:script_comparison}) reveal the advantage of the top 5 scripts: Armn, Hebr, Thai, Grek, and Cyrl.  In a twist, Latin (Latn) is not in the top nor bottom 5 scripts in terms of ChrF. The nearly identical performance patterns between XGBoost and Random Forest, combined with close alignment between predicted and actual scores, suggest that both models capture the same underlying linguistic regularities with high fidelity, making the observed categorical biases robust findings rather than model-dependent effects.

\subsection{Feature Importance Analysis}

\begin{table*}[!htbp]
\centering

\scalebox{0.9}{
\begin{tabular}{|c|c|c|c|}
\toprule
\hline
\multicolumn{2}{|c|}{\textbf{English-to-XX}} & \multicolumn{2}{c|}{\textbf{XX-to-English}} \\
\hline
\textbf{XGBoost} & \textbf{Random Forest} & \textbf{XGBoost} & \textbf{Random Forest} \\
\hline
\midrule
\small \textcolor{red}{Joshi Class} (0.365) & \small \textcolor{red}{Joshi Class} (0.205) & \small \textcolor{teal}{Region} (0.278) & \small \textcolor{teal}{Region} (0.198) \\[0.5ex]
\small \textcolor{teal}{Region} (0.206) & \small \textcolor{blue}{Language Id Code} (0.183) & \small \textcolor{orange}{Family} (0.208) & \small \textcolor{orange}{Family} (0.163) \\[0.5ex]
\small \textcolor{orange}{Family} (0.133) & \small \textcolor{teal}{Region} (0.161) & \small \textcolor{red}{Joshi Class} (0.178) & \small \textcolor{blue}{Language Id Code} (0.156) \\[0.5ex]
\small \textcolor{purple}{Script} (0.127) & \small \textcolor{green}{Reference Fertility} (0.115) & \small \textcolor{purple}{Script} (0.103) & \small \textcolor{red}{Joshi Class} (0.116) \\[0.5ex]
\small \textcolor{blue}{Language Id Code} (0.092) & \small \textcolor{magenta}{Candidate Fertility} (0.081) & \small \textcolor{blue}{Language Id Code} (0.092) & \small \textcolor{magenta}{Candidate Fertility} (0.089) \\[0.5ex]
\small \textcolor{green}{Reference Fertility} (0.025) & \small \textcolor{orange}{Family} (0.073) & \small \textcolor{green}{Reference Fertility} (0.040) & \small \textcolor{green}{Reference Fertility} (0.082) \\[0.5ex]
\small \textcolor{magenta}{Candidate Fertility} (0.021) & \small \textcolor{brown}{Reference Tokens} (0.072) & \small \textcolor{brown}{Reference Tokens} (0.041) & \small \textcolor{purple}{Script} (0.077) \\[0.5ex]
\small \textcolor{brown}{Reference Tokens} (0.018) & \small \textcolor{purple}{Script} (0.066) & \small \textcolor{magenta}{Candidate Fertility} (0.029) & \small \textcolor{brown}{Reference Tokens} (0.061) \\[0.5ex]
\small \textcolor{gray}{Candidate Tokens} (0.013) & \small \textcolor{gray}{Candidate Tokens} (0.044) & \small \textcolor{gray}{Candidate Tokens} (0.031) & \small \textcolor{gray}{Candidate Tokens} (0.059) \\
\hline
\bottomrule
\end{tabular}
}
\caption{Feature Importance Rankings by Model and Translation Direction}
\label{tab:model_direction_comparison}
\end{table*}

Table \ref{tab:model_direction_comparison} reveals distinct patterns in how XGBoost and Random Forest prioritize language-level metadata across translation directions. For English-to-XX translation, XGBoost places overwhelming emphasis on Joshi Class (0.365 importance), indicating that resource-level categorization is the most critical factor when translating into diverse target languages. This is followed by regional patterns (0.206) and language family relationships (0.133), creating a clear hierarchical structure focused on language categorizations.

Random Forest shows a more distributed approach for English-to-XX translation, with Joshi Class leading (0.205) but at much lower intensity than XGBoost, followed closely by individual language codes (0.183) and regional groupings (0.161). This more balanced distribution suggests Random Forest's ensemble approach captures a broader range of linguistic patterns rather than focusing heavily on a single dominant feature.

The XX-to-English direction shows different priorities for both models. XGBoost emphasizes regional patterns (0.278) followed by language family relationships (0.208) and Joshi Class (0.178), suggesting that when translating into English, geographic and phylogenetic groupings become more predictive than individual resource levels. Random Forest maintains region as the top feature (0.198) but shows more balanced importance across language family (0.163) and individual language codes (0.156).

The two model types diverge on fertility features.  In Random Forest, combined fertility (reference+candidate) is 0.196 for English -> XX, nearly matching Joshi Class (0.205), the most dominant feature.  XGBoost assigns fertility features lower than Random Forest, which could reflect difference in optimization strategies, where XGBoost gradient boosting prioritizes the single most discriminative splits (categorical features), while Random Forest bagging captures patterns that categorical features cannot fully represent.    

\section{Conclusion}
This work shows that much of machine translation quality can be anticipated without ever looking at the translated words themselves. Using high-level linguistic metadata, fertility ratios, and token counts, tree-based models predict ChrF scores across 200 languages with striking accuracy. While the act of predicting ChrF is not inherently remarkable, the fact that such predictions are possible without examining the text itself underscores the systematic role of fertility and typology in shaping translation quality.

Equally important, tree ensembling methods such as XGBoost not only deliver the strongest predictive performance but also provide interpretable insights into which features matter most. The relative weights assigned to fertility, resource levels, and typological categories reveal consistent cross-linguistic patterns: target-side fertility explains predictability into diverse languages, while source-side typology dominates when translating into English. These interpretable rankings allow us to see translation quality through the lens of linguistic structure rather than opaque model behavior.

Looking forward, this perspective supports the role for lightweight quality estimation as a diagnostic tool for understanding multilingual systems and the typological factors that drive their performance. By showing that translation quality can be explained largely from fertility and metadata alone, our results highlight a path toward more efficient, interpretable, and linguistically grounded approaches toward improving translation.

\section{Limitations}
Our approach to predicting translation quality from fertility and metadata, while demonstrating strong performance, has limitations. Our analysis is restricted to GPT-4o translations on the FLORES-200 benchmark, which may not generalize to other LLMs, traditional MT systems, or specialized domains. The exclusive reliance on ChrF scores, despite their correlation with human judgments, cannot capture nuanced aspects of translation quality such as cultural appropriateness or contextual accuracy. Our linguistic categorizations (family, script, region) represent coarse-grained groupings that may obscure important within-category variations, for example, treating all Niger-Congo languages uniformly ignores variation within this family. The use of a single tokenizer (o200kbase) limits our understanding of how different tokenization schemes might affect fertility-quality relationships. 
\section{Broader Impact}
This work contributes to trustworthy multilingual NLP by making translation quality disparities visible and interpretable. By identifying which linguistic factors drive performance gaps, our findings can guide targeted investment in low-resource language development and inform fairer evaluation practices.  In terms of risks, our findings reveal and potentially amplify existing biases in MT evaluation: the strong predictive power of resource-level indicators (Joshi Class) and regional groupings risks perpetuating a cycle where low-resource languages receive less attention due to anticipated lower quality scores. This could inadvertently discourage investment in improving translation for underserved languages, as stakeholders might view poor performance as inherent to these languages rather than a consequence of limited training data and research attention. Using our models for pre-deployment quality estimation could lead to discriminatory practices, such as withholding MT services from speakers of languages predicted to have lower quality, thereby exacerbating digital language divides. We therefore caution against using these predictions as gatekeeping mechanisms and instead recommend them solely as diagnostic tools for understanding systemic challenges in multilingual NLP.

\bibliography{main}




\end{document}